\documentclass[11pt,a4paper]{article}
\usepackage[nohyperref]{acl2017}

\usepackage{times}
\usepackage{url}
\usepackage{relsize} 
\usepackage{graphicx}
\usepackage{latexsym}
\usepackage{enumitem}
\usepackage{multirow}
\usepackage{subfigure}
\usepackage{amsmath}
\usepackage{comment}
\usepackage{color}
\usepackage{array}
\newcolumntype{P}[1]{>{\centering\arraybackslash}p{#1}}
\usepackage[font=small,labelfont=bf]{caption}
\DeclareCaptionFont{tiny}{\tiny}
\newcommand{\todo}[1]{\textcolor{red}{TODO: #1}}

\newcommand\Mark[1]{\textsuperscript#1}

\aclfinalcopy

\title{Adapting predominant and novel sense discovery algorithms for identifying corpus-specific sense differences}

\author{Binny Mathew\Mark{1}, Suman Kalyan Maity\Mark{2}, Pratip Sarkar\Mark{3}, Animesh Mukherjee\Mark{4} and Pawan Goyal\Mark{5} \\
Department of Computer Science and Engineering \\
Indian Institute of Technology Kharagpur, India - 721302 \\
Email: \{binny.iitkgp\Mark{1}, pratip.sarkar.iitkgp\Mark{3}, animeshm\Mark{4}, pawang.iitk\Mark{5}\}@gmail.com \\ sumankalyan.maity@cse.iitkgp.ernet.in\Mark{2}
}
\date{08-07-2016}
\begin{document}
\maketitle

\begin{abstract}
Word senses are not static and may have temporal, spatial or corpus-specific scopes. Identifying such scopes might benefit the existing WSD systems largely. In this paper, while studying corpus specific word senses, we adapt three existing predominant and novel-sense discovery algorithms to identify these corpus-specific senses. We make use of text data available in the form of millions of digitized books and newspaper archives as two different sources of corpora and propose automated methods to identify corpus-specific word senses at various time points. We conduct an extensive and thorough human judgment experiment to rigorously evaluate and compare the performance of these approaches. Post adaptation, the output of the three algorithms are in the same format and the accuracy results are also comparable, with roughly {\bf 45-60\%} of the reported corpus-specific senses being judged as genuine.

\end{abstract}
\vspace{-0.3cm}
\section{Introduction}
\label{sec:intro}
\vspace{-0.2cm}
Human language is neither static not uniform. Almost every individual aspect of language including phonological, morphological, syntactic as well as semantic structure can exhibit differences, even for the same language. These differences can be influenced by a lot of factors such as time, location, corpus type etc. However, in order to suitably understand these differences, one needs to be able to analyze large volumes of natural language text data collected from diverse corpora. It is only in this Big Data era that unprecedented amounts of text data have become available in the form of millions of digitized books (Google Books project), newspaper documents, Wikipedia articles as well as tweet streams. This huge volume of time and location stamped data across various types of corpora now allows us to make precise quantitative linguistic predictions, which were earlier observed only through mathematical models and computer simulations. 

\noindent{\bf Scope of a word sense:} One of the fundamental dimensions of language change is shift in word usage and word senses~\cite{Jones:86,Ide:98,Schutze:98,Navigli:09}. A word may possess many senses; however, not all of the senses are used uniformly; some are more common than the others. This particular distribution can be heavily dependent on the underlying time-period, location or the type of corpora. For example, let us consider the word ``rock''. In books, it is usually associated with the sense reflected by the words `stone, pebble, boulder' etc., while if we look into newspapers and magazines, we find that it is mostly used in the sense of `rock music'.


\begin{figure*}[!ht]
\centering
  \begin{tabular}{@{}cc@{}}
    \includegraphics[width=.45\textwidth]{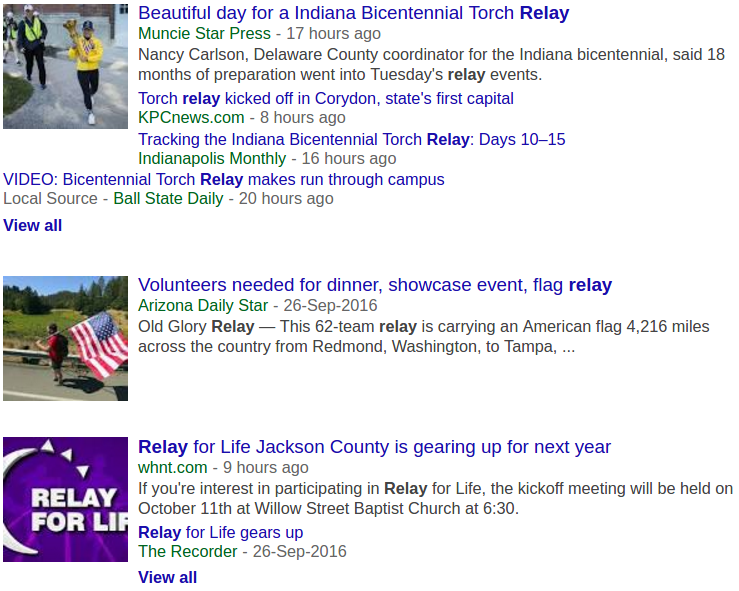} &
    \includegraphics[width=.45\textwidth]{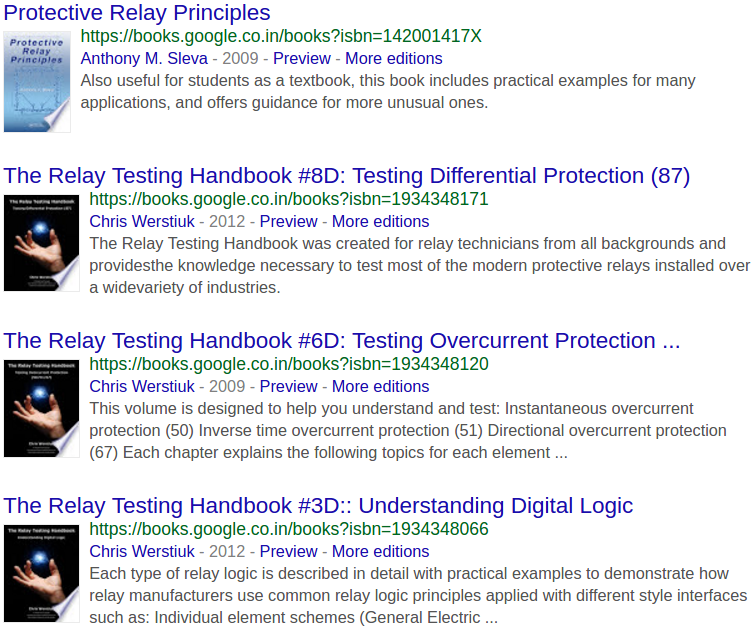} 
       \\
       (a) & (b)
  \end{tabular}
  \caption{Google search results for the word 'relay' using (a) Google News and (b) Google Books. }
  \label{fig:relay_motivation}
\end{figure*}

\noindent{\bf Motivation for this work:} The world of technology is changing rapidly, and it is no surprise that word senses also reflect this change. Let us consider the word ``brand''. This word is mainly used for the `brand-name' of a product. However, it has now become a shorthand reference to the skills, actions, personality and other publicly perceived traits of individuals or for characterizing reputation, public face of the whole group or companies. The rise of social media and the ability to self-publish and self-advertise undoubtedly led to the emergence of this new sense of ``brand''. To further motivate such cross corpus sense differences, let us consider the word 'relay'. A simple Google search in the News section produces results that are very different from those obtained through a search in the Books section (See Fig ~\ref{fig:relay_motivation}). In this paper, we attempt to automatically build corpus-specific contexts of a target word (for e.g., relay in this case) that can appropriately discriminate the two different senses of the target word -- one of which is more relevant for the News corpus (context words extracted by one of our adapted methods: \textit{team, race, event, races, sprint, men, events, record, run, win}) while the other is more relevant for the Books corpus (context words extracted by one of our adapted methods: \textit{solenoid, transformer, circuitry, generator, diode, sensor, transistor, converter, capacitor, transformers}). 
Since the search engine users mostly go for generic search without any explicit mention of book or news, the target word along with a small associated context vector might help the search engine to retrieve document from the most relevant corpora automatically. We believe that the target and the automatically extracted corpus-specific context vector can be further used to enhance (i) semantic and personalized search, (ii) corpora-specific search and (iii) corpora-specific word sense disambiguation. It is an important as well as challenging task to identify predominant word senses specific to various corpora. While the researchers have started exploring the temporal and spatial scopes of word senses~\cite{Cook:10,Gulordava:11,Kulkarni:15,Jatowt:14,Mitra:14,Mitra:15}, corpora-specific senses have remained mostly unexplored. 



\noindent{\bf Our contributions:} Motivated by the above applications, this paper studies corpora-specific senses for the first time and makes the following contributions \footnote{The code used to find corpus-specific senses and evaluation results are available at: \url{http://tinyurl.com/h4onyww}}: (i) we take two different methods for novel sense discovery~\cite{Mitra:14,Lau:2014} and one for predominant sense identification~\cite{McCarthy:2004} and adapt these in an automated and unsupervised manner to identify corpus-specific sense for a given word (noun), and (ii) perform a thorough manual evaluation to rigorously compare the corpus-specific senses obtained using these methods. Manual evaluation conducted using $60$ candidate words for each method indicates that $\sim${\bf 45-60\%} of the corpus-specific senses identified by the adapted algorithms are genuine. Our work is a unique contribution since it is able to adapt three very different types of major algorithms suitably to identify corpora specific senses.

\noindent{\bf Key observations:} For manual evaluation of the candidate corpus-specific senses, we focused on two aspects -- a) {\sl sense representation}, which tells if the word cluster obtained from a method is a good representative of the target word, and b) {\sl sense difference}, which tells whether the sense represented by the corpus-specific cluster is different from all the senses of the word in the other corpus. Some of our important findings from this study are: (i) the number of candidate senses produced by McCarthy \textit{et al.}~\shortcite{McCarthy:2004} is far less than the two other methods, (ii) Mitra \textit{et al.}~\shortcite{Mitra:14} produces the best representative sense cluster for a word in the time period 2006-2008 and McCarthy \textit{et al.}~\shortcite{McCarthy:2004} produces the best representative sense cluster for a word  in the time period 1987-1995, (iii) Mitra \textit{et al.}~\shortcite{Mitra:14} is able to identify sense differences more accurately in comparison to the other methods, (iv) considering both the aspects together, McCarthy \textit{et al.}~\shortcite{McCarthy:2004} performs the best, (v) for the common results produced by Lau \textit{et al.}~\shortcite{Lau:2014} and Mitra \textit{et al.}~\shortcite{Mitra:14}, the former does better sense differentiation while the latter does better overall. 
\section{Related Work}
\label{sec:related}
\vspace{-0.2cm}
Automatic discovery and disambiguation of word senses from a given text is an important and challenging problem, which has been extensively studied in the literature~\cite{Jones:86,Ide:98,Schutze:98,Navigli:09,Kilgarriff:01,Kilgarriff:04}. Only recently, with the availability of enormous amounts of data, researchers are exploring temporal scopes of word senses. Cook and Stevenson~\shortcite{Cook:10} use corpora from different time periods to study the change in the semantic orientation of words. Gulordava and Baroni~\shortcite{Gulordava:11} use two different time periods in the Google n-grams corpus and detect semantic change based on distributional similarity between word vectors. 
Kulkarni \textit{et al.}~\shortcite{Kulkarni:15} propose a computation model for tracking and detecting statistically significant linguistic shifts in the meaning and usage of words. Jatowt and Duh~\shortcite{Jatowt:14} propose a framework for exploring semantic change of words over time on Google n-grams and COHA dataset. Lau \textit{et al.}~\shortcite{Lau:2014} propose a fully unsupervised topic modelling-based approach to sense frequency estimation, which was used for the tasks of predominant sense learning, sense distribution acquisition, detecting senses which are not attested in the corpus, and identifying novel senses in the corpus which are not captured in the sense inventory. Two recent studies by Mitra \textit{et al.}~\shortcite{Mitra:14,Mitra:15} capture temporal noun sense changes by proposing a graph clustering based framework for analysis of diachronic text data available from Google books as well as tweets. 
quantify semantic change by evaluating word embeddings against known historical changes. Lea and Mirella ~\shortcite{Lea:2016} develop a dynamic Bayesian model of diachronic meaning change. Pelevina ~\shortcite{Pelevina:16} develops an approach which induces a sense inventory from existing word embeddings via clustering of ego-networks of related words.

Cook \textit{et al.}~\shortcite{Cook:13} induce word senses and then identify novel senses by comparing two different corpora: the `focus corpora' (i.e., a recent version of the corpora) and the `reference corpora' (older version of the corpora). Tahmasebi \textit{et al.}~\shortcite{Tahmasebi:11}, propose a framework for tracking senses in a newspaper corpus containing articles between 1785 and 1985. Phani \textit{et al.}~\shortcite{Phani:12} study 11 years worth Bengali newswire that allows them to extract trajectories of salient words that are of importance in contemporary West Bengal. Few works ~\cite{Dorow:03,McCarthy:2004} have focused on corpus-specific sense identification. 
Our work differs from these works in that we capture the cross corpus-specific sense differences by comparing the senses of a particular word obtained across two different corpora. We adapt three state-of-the-art novel and predominant sense discovery algorithms and extensively compare their performances for this task.  
%
\section{Dataset Description}
\label{sec:dataset}
To study corpora-specific senses, we consider books and newspaper articles as two different corpora sources. We compare these corpora for the same time-periods to ensure that the sense differences are obtained only because of the change in corpus and not due to the difference in time. A brief description of these datasets is given below.\\
\textbf{Books dataset:} The books dataset is based on the Google Books Syntactic n-grams corpus~\cite{Goldberg:13}, consisting of time-stamped texts from over 3.4 million digitized English books, published between 1520 and 2008. For our study, we consider Google books data for the two time periods $1987-1995$ and $2006-2008$.\\ 
\textbf{Newspaper dataset:} For the Newspaper dataset, we consider two different data sources. The first dataset from $1987-1995$ contains articles of various newspapers\footnote{https://catalog.ldc.upenn.edu/LDC93T3A}. The other dataset from $2006-2008$ is gathered from the archives of The New York Times.

\section{Proposed framework}
\label{sec:framework}
To identify corpus-specific word senses, we aim at adapting some of the existing algorithms, which have been utilized for related tasks. In principle, we compare all the senses of a word in one corpus against all the senses of the same word in another corpus. We, therefore, base this work on three different approaches, Mitra \textit{et al.}~\shortcite{Mitra:14}, Lau \textit{et al.}~\shortcite{Lau:2014} and McCarthy \textit{et al.}~\shortcite{McCarthy:2004}, which could be adapted to find word senses in different corpora in an unsupervised manner. Next, we discuss these methods briefly followed by the proposed adaptation technique and generation of the candidate set.

\begin{figure*}[htb]
\centering
  \begin{tabular}{@{}ccc@{}}
    \includegraphics[width=.3\textwidth]{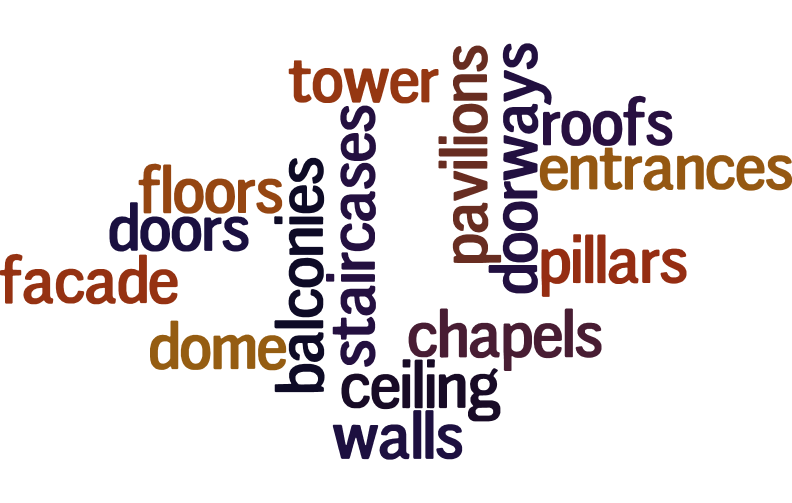} &
    \includegraphics[width=.3\textwidth]{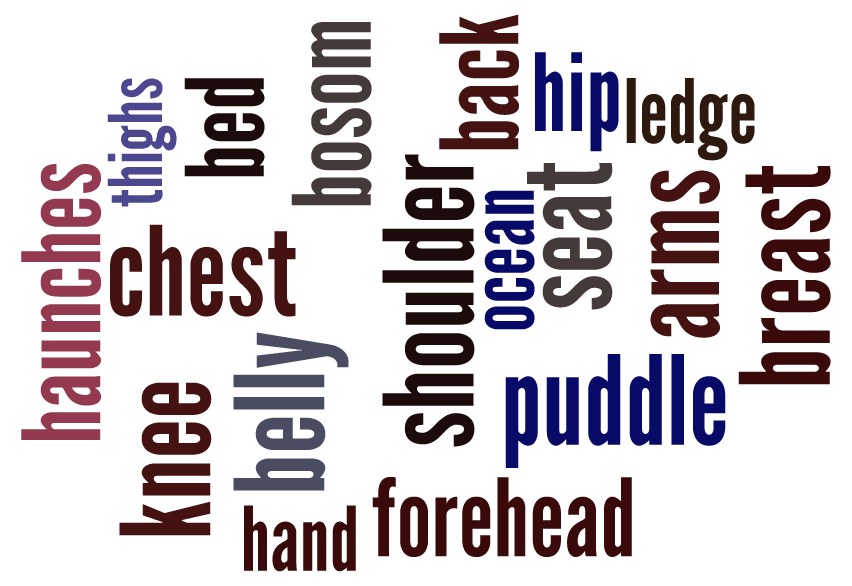} &
    \includegraphics[width=.3\textwidth]{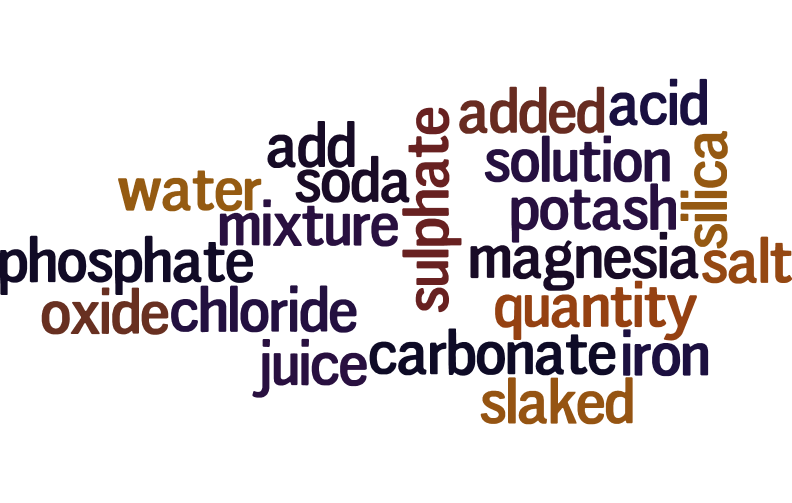} 
       \\
       (a) & (c) & (e)\\
    \includegraphics[width=.3\textwidth]{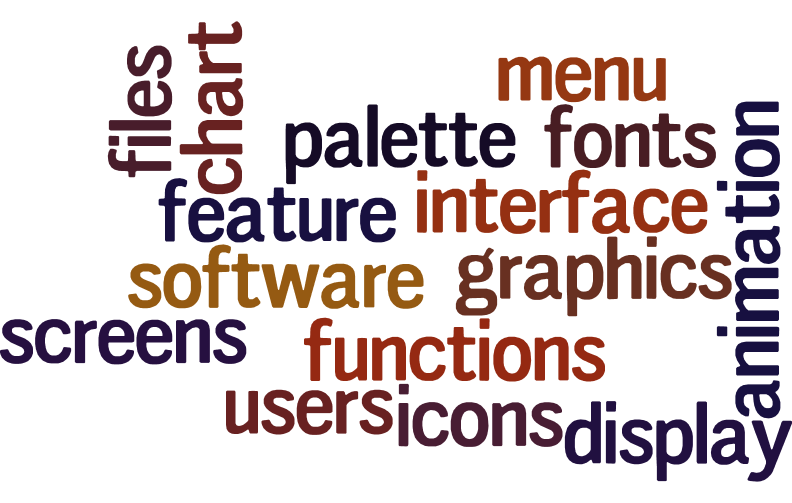} &
    \includegraphics[width=.3\textwidth]{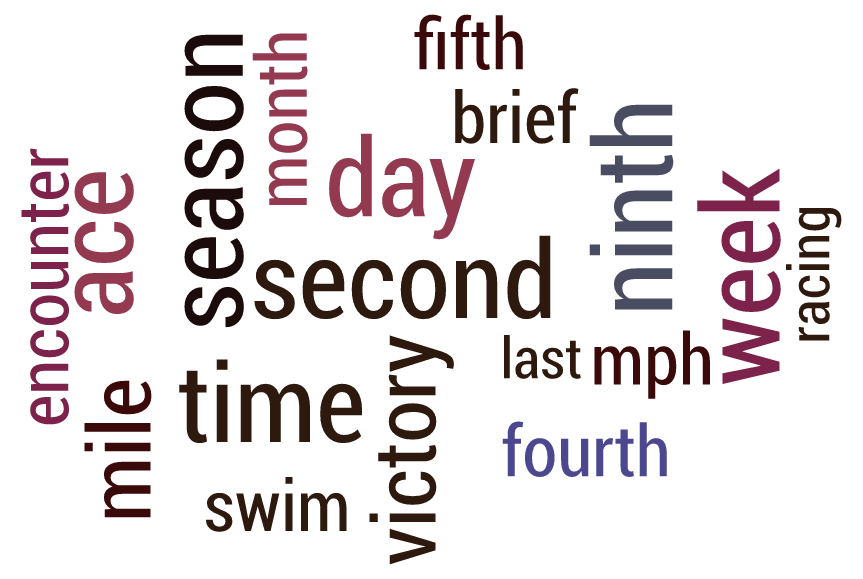} &
    \includegraphics[width=.3\textwidth]{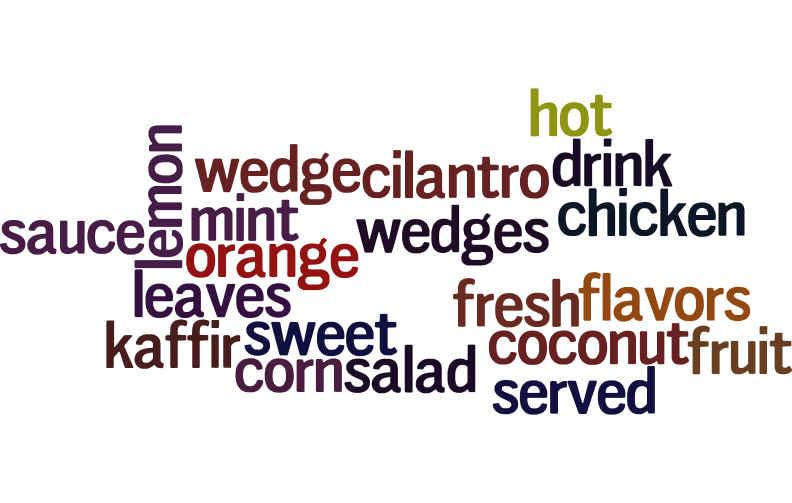} \\
    (b) & (d) & (f)
  \end{tabular}
  \caption{Examples of corpora-specific sense clusters obtained for (a,b) `windows' using Mitra's method for (books, news) during 1987-1995, (c,d) `lap' using McCarthy's method for (books, news) during 2006-2008 and (e,f) `lime' using Lau's method for (books, news) during 2006-2008.}
  \label{fig:birth}
\end{figure*}

\vspace{-0.0cm}
\subsection{Mitra's Method}
Mitra \textit{et al.}~\shortcite{Mitra:14} proposed an unsupervised method to identify noun sense changes over time. They prepare separate distributional-thesaurus-based networks (DT)~\cite{Riedl:13} for the two different time periods. Once the DTs have been constructed, Chinese Whispers (CW) algorithm~\cite{Biemann:06} is used for inducing word senses over each DT. For a given word, the sense clusters across two time-points are compared using a split-join algorithm.

\noindent {\bf Proposed adaptation:} In our adaptation, we apply the same framework but over the two different corpora sources in the same time period. So, for a given word $w$ that appears in both the books and newspaper datasets, we get two different set of clusters, $B$ and $N$, respectively for the two datasets. Accordingly, let $B=\{s_{b1},s_{b2},\ldots,s_{b|B|}\}$ and $N=\{s_{n1},s_{n2},\ldots,s_{n|N|}\}$, where $s_{bi}$ ($s_{nj}$) denotes a sense cluster for $w$ in the books (news) dataset. 

A corpus-specific sense will predominantly be present only in that specific corpus and will be absent from the other corpus. To detect the book-specific sense for the word $w$, we compare each of the $|B|$ book clusters against all of the $|N|$ newspaper clusters. Thus, for each cluster $s_{bi}$, we identify the fraction of words that are not present in any of the $|N|$ newspaper clusters. If this value is above a threshold, we call $s_{bi}$ a book-specific sense cluster for the word $w$. This threshold has been set to $0.8$ for all the experiments, as also reported in Mitra \textit{et al.}~\shortcite{Mitra:14}.

We also apply the multi-stage filtering\footnote{majority voting after multiple runs of CW and POS tags `NN' and `NNS'} to obtain the candidate words as mentioned in their paper, except that we do not filter the top $20\%$ and bottom $20\%$ of the words. We believe that removing the top $20\%$ words would deprive us of many good cases. To take care of the rare words, we consider only those corpus-specific clusters that have $\geq10$ words .
The number of candidate words obtained after this filtering are shown in Table \ref{tab:wn_filtered}. Figure~\ref{fig:birth} (a,b) illustrates two different sense clusters of the word `windows' - one specific to books corpus and another specific to newspaper corpus, as obtained using Mitra's method. The book-specific sense corresponds to `an opening in the wall or roof of a building'. The newspaper-specific sense, on the other hand, is related to the computing domain, suggesting Windows operating system.

\begin{table}[thb]
\vspace{-0.0cm}
\centering
\begin{small}
\caption{Number of candidate corpus-specific senses using Mitra's method after multi-stage filtering}
\begin{tabular}{|c|c|c|}
\hline
           & 1987-1995 & 2006-2008 \\ \hline
Books      & 32036      & 30396      \\ \hline
Newspapers & 18693      & 20896      \\ \hline
\end{tabular}
\label{tab:wn_filtered}
\end{small}
\vspace{-0.0cm}
\end{table}

\subsection{McCarthy's Method}
McCarthy et. al.~\shortcite{McCarthy:2004} developed a method to find the predominant sense of target word $w$ in a given corpora. The method requires the nearest neighbors to the target word, along with the distributional similarity score between the target word and its neighbors. It then assigns a prevalence score to each of the WordNet synset $ws_i$ of $w$ by comparing this synset to the neighbors of $w$. The prevalence score $PS_i$ for the synset $ws_i$ is given by

\begin{small}
\vspace{-0.3cm}
\begin{equation}
PS_i=
\sum\limits_{n_{j} \in N_{w}}dss(w,n_{j}) \times \frac{wnss(ws_{i},n_{j})}{\sum\limits_{ws_{i'}}wnss(ws_{i'},n_{j})}
\vspace{-0.3cm}
\end{equation}
\end{small}
where $N_w$ denotes the set of neighbors of $w$ and $dss(w,n_j)$ denotes the distributional similarity between word $w$ and its neighbors $n_j$. $wnss(ws_i,n_j)$ denotes the WordNet similarity between the synset $ws_i$ and the word $n_j$, and is given by
\begin{equation}
wnss(ws_{i},n_{j})=\max\limits_{ns_{x} \in senses(n_{j})}ss(ws_{i},ns_{x})
\end{equation}
where $ss(ws_{i},ns_{x})$ denotes the semantic similarity between WordNet synsets $ws_i$ and $ns_x$. We use Lin Similarity measure to find similarity between two WordNet synsets.

\noindent{\bf Proposed adaptation}: In our adaptation to McCarthy's method to find corpus-specific senses, we use the DT networks constructed for Mitra's method to obtain the neighbors as well as distributional similarity between a word and its neighbors. We then obtain the prevalence score for each sense of the target word for both the corpora sources separately, and normalize these scores so that the scores add up to 1.0 for each corpus. We call these as normalized prevalence score ($NPS$).

We call a sense $ws_i$ as corpora specific if its $NPS_i$ is greater than an upper threshold in one corpus and less than a lower threshold in the other corpus. We use 0.4 as the upper threshold and 0.1 as the lower threshold for our experiments. After applying this threshold, the number of candidate words are shown in Table \ref{tab:McCarthy_words}.

\begin{table}[thb]
\vspace{-0.0cm}
\centering
\begin{small}
\caption{Number of candidate corpus-specific senses using McCarthy's method.}
\begin{tabular}{|c|c|c|}
\hline
           & 1987-1995 & 2006-2008 \\ \hline
Books      & 97       & 95       \\ \hline
Newspapers & 117       & 97       \\ \hline
\end{tabular}
\label{tab:McCarthy_words}
\end{small}
\end{table}

For the purpose of distributional visualization of the senses, we denote a word sense $ws_i$ using those neighbors of the word, which make the highest contribution to the prevalence score $PS_i$. Figure ~\ref{fig:birth} (c, d) illustrates two sense clusters of the word `lap' thus obtained - one specific to books corpus and another specific to newspaper corpus. The book-specific sense corresponds to `the top surface of the upper part of the legs of a person who is sitting down'. The news-specific sense, on the other hand corresponds to `a complete trip around a race track that is repeated several times during a competition'. 

\subsection{Lau's Method}

We also adapt the method described in Lau \textit{et al.}~\shortcite{Lau:2014} to find corpus specific word senses. Their method uses topic modeling to estimate word sense distributions and is based on the word sense induction (WSI) system described in Lau \textit{et al.}~\shortcite{Lau:2012}. The system is built around a Hierarchical Dirichlet Process (HDP)~\cite{Teh:2006}, which optimises the number of topics in a fully-unsupervised fashion over the training data. For each word, they first induce topics using HDP. The words having the highest probabilities in each topic denote the sense cluster. The authors treat the novel sense identification task as identifying sense clusters that do not align well with any of the pre-existing senses in the sense inventory. They use topic-to-sense affinity to estimate the similarity of a topic to the set of senses given as
\begin{equation}
ts-affinity(t_j)=\frac{\sum_{i}^{S} Sim(s_{i},t_{j})}{\sum_{l}^{T} \sum_{k}^{S} Sim(s_{k},t_{l})}
\vspace{-0.2cm}
\end{equation}

where $T$ and $S$ represent the number of topics and senses respectively, and $Sim(s_{i},t_{j})$ is defined as
\begin{equation}
Sim(s_{i},t_{j})=1 - JS(S_{i} || T_{j})
\label{equ:sim}
\vspace{-0.2cm}
\end{equation}
where $S_i$ and $T_j$ denote the multinomial distributions over words for sense $s_i$ and topic $t_j$. $JS(X,Y)$ stands for Jensen-Shannon divergence between distributions $X$ and $Y$.



\noindent{\bf Proposed adaptation:} In our adaptation to their method to find corpus-specific senses, for a target word, a topic is called corpus-specific if its word distributions are very different from all the topics in the other corpus. We therefore compute similarity of this topic to all the topics in other corpus and if the maximum similarity is below a threshold, this topic is called as corpus-specific. We use Equation~\ref{equ:sim} to compute the similarity between two topics $t_i$ and $t_j$ as $Sim(t_i,t_j)$.



Since Lau's method is computationally expensive to run over the whole vocabulary, we run it only for those candidate words, which were flagged by Mitra's method. 
We then use a threshold to select only those topics which have low similarity to all the topics in the other corpus. We use 0.35 as the threshold for all the 4 cases except for news-specific senses in 2006-2008, where a threshold of 0.2 was used. The number of candidate corpus-specific senses thus obtained are shown in Table~\ref{tab:Cooks_words}. Note that a word may have multiple corpus-specific senses. 

\begin{table}[thb]
\vspace{-0.1cm}
\centering
\begin{small}
\caption{Number of candidate words using Lau's method.}
\begin{tabular}{|c|c|c|}
\hline
           & 1987-1995 & 2006-2008 \\ \hline
Books      & 6478       & 4339       \\ \hline
Newspapers & 23587       & 1944       \\ \hline
\end{tabular}
\label{tab:Cooks_words}
\end{small}
\vspace{-0.1cm}
\end{table}

Figure ~\ref{fig:birth}(e,f) illustrates the two different word clusters of the word `lime' - one specific to the books corpus and another specific to the newspaper corpus, as obtained by applying their method. The book-specific sense corresponds to `mineral and industrial forms of calcium oxide'. The news-specific sense, on the other hand, is related to `lemon, lime juice'.

\begin{table*}[!thb]
\begin{small}
\begin{center}
\captionsetup{font=small}
\caption{Accuracy figures for the three methods from manual evaluation.}
\resizebox{\textwidth}{!}{%
\begin{tabular}
{|c|c|c|c|c|c|c|c|c|}\hline
& & \multicolumn{2}{c|}{Sense Representation} & \multicolumn{3}{c|}{Sense Discrimination} & \multicolumn{2}{c|}{Overall Confidence} \\ \hline
Method & Time-period & Majority voting & Average & Majority voting& Average & Undecided & Majority voting & Average \\ \hline
\multirow{2}{*}{Lau} & 1987-1995 & 46.67\% & 60.0\% & 40.0\% & 61.82\% & 33.33\% & 30.0\% & 37.78\% \\ \cline{2-9}
     & 2006-2008 & 70.0\% & 67.78\% & 50.0\% & \textbf{63.93\%} &  23.33\%&43.33\% & 44.44\%\\ \hline
\multirow{2}{*}{McCarthy} & 1987-1995 & \textbf{76.67\%} & \textbf{77.78\%} & 66.67\% & \textbf{78.57\%} & 20.0\% & \textbf{56.67\%} & \textbf{61.11\%} \\ \cline{2-9}
         & 2006-2008 & 66.67\% & 68.89\% & 53.33\% & 55.0\% & 6.67\% & \textbf{46.67\%} & \textbf{48.89\%} \\ \hline
\multirow{2}{*}{Mitra} & 1987-1995 & 75.0\% & 76.19\% & \textbf{73.91\%} & 66.2\% & \textbf{17.86\%} & 50.0\% & 50.0\% \\ \cline{2-9}
         & 2006-2008 & \textbf{87.5\%} & \textbf{80.21\%} & \textbf{60.0\%} & 57.47\% & \textbf{6.25\%} & 44.79\% & 46.88\%\\ \hline
\end{tabular}%
}
\label{tab:sense_combined}
\end{center}
\end{small}
\end{table*}

\begin{table*}[!thb]
\begin{small}
\begin{center}
\captionsetup{font=small}
\caption{Comparison of accuracy figures for 30 overlap words between Lau and Mitra.}
\begin{tabular}
{|c|c|c|c|c|c|c|c|}\hline
& \multicolumn{2}{c|}{Sense Representation} & \multicolumn{3}{c|}{Sense Discrimination} & \multicolumn{2}{c|}{Overall Confidence} \\ \hline
Method & Majority voting & Average & Majority voting & Average & Undecided & Majority voting & Average \\ \hline
Lau &  50.0\% & 53.33\% & \textbf{65.38\%} & \textbf{55.56\%} & 13.33\% & 26.67\% & 26.67\% \\ \hline
Mitra & \textbf{90.0\%} & \textbf{84.44\%} & 50.0\% & 48.89\% & 13.33\% & \textbf{41.11\%} & \textbf{43.33\%}   \\ \hline
\end{tabular}
\label{tab:overlap_combined}
\end{center}
\end{small}
\end{table*}

\begin{table}[thb]
\centering
\begin{small}
\caption{Fleiss' kappa for the three methods}
\begin{tabular}{|c|c|c|c|}
\hline
    & Lau & McCarthy & Mitra \\ \hline
Question 1     & 0.40       & 0.31 & 0.41      \\ \hline
Question 2 & 0.19       & 0.12 & 0.12      \\ \hline
\end{tabular}
\label{tab:Fleiss_kappa}
\end{small}
\end{table}

\section{Evaluation Framework and Results}
\label{sec:evaluation}
\vspace{-0.2cm}
In this section, we discuss our framework for evaluating the candidate corpus-specific senses obtained from the three methods. We perform manual evaluations using an online survey\footnote{\url{http://tinyurl.com/zd2hmef}} among $\sim27$ agreed participants (students, researchers, professors, technical persons) with age between 18-34 years. 
We randomly selected 60 candidate corpus-specific senses (combining both corpora sources) from each of the three methods (roughly 30 words from each time period). Each participant was given a set of 20 candidate words to evaluate; thus each candidate sense was evaluated by 3 different annotators. In the survey, the candidate word was provided with its corpus-specific sense cluster (represented by word-clouds of the words in the cluster) and all the sense clusters in the other corpus.

\noindent {\bf Questions to the participants:} The participants were asked two questions. First, {\sl whether the candidate corpus-specific sense cluster is a good representative sense of the target word?} and second, {\sl whether the sense represented by the corpus-specific cluster is different from all the senses of the word in the other corpus?} The participants could answer the first question as `Yes' or `No' and this response was taken as a measure of ``sense representation" accuracy of the underlying scheme. If this answer is `No', the answer to the second response was set as `NA'. If this answer is `Yes', they would answer the second question as `Yes' or `No', which was taken as a measure of ``discriminative sense detection" accuracy of the underlying method for comparing the senses across the two corpora. The overall confidence of a method was obtained by combining the two responses, i.e., whether both the responses are `Yes'. The accuracy values are computed using {\sl majority voting}, where we take the output as `Yes' if majority of the responses are in agreement with the system and {\sl average accuracy}, where we find the fraction of responses that are in agreement with the system. Since each case is evaluated by 3 participants, micro- and macro-averages will be similar.

\noindent{\bf Accuracy results:} Table~\ref{tab:sense_combined} shows the accuracy figures for the underlying methods. Mitra's and McCarthy's methods perform better for sense representation, and Mitra's method performs very well for discriminative sense detection. For discriminative sense detection, there were a few undecided cases\footnote{This happens when one of the three annotators responded the first question as `No', thus leaving only two valid responses for the second question. If both responses do not match, majority voting will remain undecided.}. 
As per overall confidence, we observe that McCarthy's method performs the best. Note that the number of candidate senses returned by McCarthy were much less in comparison to the other methods. Mitra's method performs comparably for both the time periods, while Lau's method performs comparably only for 2006-2008. 

\noindent{\bf Inter-annotator agreement:} The inter-annotator agreement for the three methods using Fleiss' kappa is shown in Table~\ref{tab:Fleiss_kappa}. We see that the inter-annotator agreement for Question 2 is much less in comparison to that for Question 1. This is quite natural since Question 2 is much more difficult to answer than Question 1 even for humans. 

\begin{table*}[!thb]
\large
\begin{center}
\caption{Example cases from the evaluation: First column mentions the method name, which corpus-specific, time-period and the candidate word. Second column mentions the responses to the two questions. Corpus-specific sense cluster is shown in third column and fourth column shows the sense clusters in the other corpus, separated by `\#\#'.}
\resizebox{1.00\textwidth}{!}{%
\begin{tabular}
{|p{2.0cm}|p{1.5cm}|p{8cm}|p{15cm}|}\hline
Description  & Response & \centering Corpus-specific sense cluster &   \parbox{12cm}{ \centering Sense clusters in other corpus}  \\ \hline
 Lau,   \textbf{News},  \textbf{2006-2008},   navigation & Yes, Yes & devices, gps, systems, company, mobile, portable, device, software, oriental, steam, co., peninsular, market, personal, products, ports, tomtom, car, digital, \ldots & company, river, commerce, steam, act, system, free, mississippi, \ldots \#\# spend, academic, according, activities, age, area, artistic, athletic, \ldots \#\# engaged, devoted, literary, agricultural, intellectual, devote, interest, occupied, \ldots \#\# pleasures, nature, mind, literature, amusements, \ldots  \\ \hline
     Lau, \textbf{Book},  \textbf{2006-2008},  fencing & Yes, No & riding, dancing, taught, exercises, boxing, drawing, horses, archery, study, horsemanship, music, swimming, wrestling, schools, \ldots & team, club, olympic, school, women, sport, sports, gold, \ldots \#\# border, miles, barriers, build, billion, congress, bill, illegal, \ldots \#\# security, wire, area, park, construction, fence, property, city,  \ldots \\ \hline
     Lau, \textbf{Book},  \textbf{1987-1995},  stalemate & No, NA & york, break, hansen, south, front, hill, turned, bloody, north, western, provide, knopf, talbott, breaking, \ldots & political, government, minister, president, prime, opposition, coalition, aimed, \ldots \#\# budget, house, congress, federal, tax, bush, white, senate, \ldots \#\# war, military, ended, president, states, talks, peace, conflict, \ldots \\ \hline
\multirow{1}{*}{McCarthy},  \textbf{Book},  \textbf{2006-2008},  pisces & Yes , Yes & scorpio, aquarius, libra, aries, sagittarius, leo, cancer, constellation, constellations, orion, capricornus, scorpius, perseus, uranus, pluto, auriga, andromeda, bootes, ophiuchus, \ldots & protocol, putt, shootings, aspect, golf, yes, relationships, onset, \ldots \#\# tablets, economist, guides, realist, officer, attorney, trustees, chairmen, \ldots \#\# hearings, bottom, peak, surface, floors, floor, walls, berm, \ldots \\ \hline
     McCarthy, \textbf{News},  \textbf{2006-2008},  filibuster & Yes, No & rebellion, insurgency, combat, decision, campaign, crackdown, determination, objections, crusade, amendments, offensive, wars, interference, assault, violation, battle, dishonesty, \ldots & pirates, raiders, invaders, adventurers, bandits, smugglers, freebooters, privateers, vikings, robbers, corsairs, outlaws, buccaneers, rebels, traders, marauders, tribesmen, brigands, slavers, insurgents,  \ldots  \\ \hline
     McCarthy, \textbf{News},  \textbf{1987-1995},  agora & No, NA & opinions, restriction, appetite, rubric, pandions, authorizations, nato, delegations, bannockburn, dm, ceding, resolve, industrialization, cry, miracle, gop, shortage, navy, yes, multimedia, \ldots & marketplace, plaza, courtyard, acropolis, stadium, precinct, sanctuary, pompeii, piazza, auditorium, temple, synagogues, basilica, synagogue, cemeteries, arena, gymnasium, palace, portico, amphitheatre,  \ldots  \\ \hline
\multirow{1}{*}{Mitra},  \textbf{News},  \textbf{2006-2008},  chain & Yes, Yes & carrier, empire, business, retailer, bank, supplier, franchise, franchises, corporation, firms, brands, distributor, firm, seller, group, organization, lender, conglomerate, provider, businesses, manufacturer, giant, company,  \ldots & fiber, filament, polymer, hydrocarbon, \ldots \#\# network, mesh, lattice, \ldots \#\# ladder, hierarchy, \ldots \#\# subunit, molecules, protein, macromolecules, molecule, subunits, receptor, chains, \ldots \#\# bracelet, necklaces, earrings, brooch, necklace, bracelets, pendant, rosary, \ldots \#\# pin, knot, noose, girdle, knob, scarf, leash, pulley, \ldots \#\# bond, bonds, \ldots \#\# never, still, fast, \ldots \#\# non, \ldots \#\# proton, \ldots \#\# test, four, per, triple, ten, multi, two, square \ldots \#\# air, neck, computer, under, cigar, bank, load, pressure, \ldots \\ \hline
     Mitra, \textbf{Book},  \textbf{1987-1995},  divider & Yes, No & potentiometer, voltmeter, oscilloscope, converters, oscillator, connector, amplifier, filtering, coupler, filter, microphone, accelerator, reflector, relay, signal, probe, regulator, preamplifier, oscillators, array, multiplier, \ldots & pulses, amplifiers, proportional, pulse, signal, frequencies, amplifier, voltage, \ldots \#\# chip, circuits, circuitry, clock, arrays, \ldots \#\# chambers, wall, junction, openings, barriers, dividers, semiconductor, wires, \ldots \#\# below, level, above, deviation, \ldots \#\# truck, planes, plane, van, motorists, lanes, \ldots \#\# addresses, \ldots \#\# along, gate, stone, gates, fence, \ldots \#\# modes, widths, rotation, projection, form, densities, model \ldots  \\ \hline
     Mitra, \textbf{News},  \textbf{1987-1995},  explanations & No, NA & way, qualities, phrases, indications, impression, manner, experience, wisdom, assumption, view, judgments, rumors, sentences, \ldots & causes, evidence, \ldots \#\# theses, motivations, judgements, analyses, inferences, answers, definitions, predictions, \ldots \#\# proxy, blame, accounting, reasons, accounting, blamed, remedies, compensates,  \ldots \\ \hline
\end{tabular}%
}
\label{tab:response_example}
\end{center}
\end{table*}

\noindent{\bf Comparison among methods:} Further, we wanted to check the relative performance of the three approaches on a common set of words. McCarthy's output did not have any overlap with the other methods but for Lau and Mitra, among the words selected for manual evaluation, 30 words were common. We show the comparison results in Table ~\ref{tab:overlap_combined}. While Lau performs better on discriminative sense detection accuracy, Mitra performs much better overall.

\section{Discussion}
\label{sec:discussion}

In this section, we discuss the results further by analyzing some of the responses. In Table \ref{tab:response_example}, we provide one example entry each for all the three possible responses for the three methods. 

\begin{table*}[!thb]
\centering
\begin{small}
\caption{Results for different thresholds of McCarthy's method to make a total of 50 words. Each cell represents the total number of words (number of candidate words chosen for a threshold + number of candidate words from the previous thresholds = total number of candidate words) (overall confidence).}
\begin{tabular}{|c|c|c|c|c|}
\cline{3-5}
\multicolumn{2}{c|}{}    & \multicolumn{3}{c|}{Upper Threshold} \\ \cline{3-5}
\multicolumn{2}{c|}{}    & 0.45 & 0.40 & 0.35 \\ \hline
\multirow{3}{*}{\parbox{2cm}{\centering Lower Threshold}} & 0.05     & 69 (2) (50\%)       & 105 ($2+(2)$) (50\%) & 152 ($2+(4)$) (33.33\%)      \\ \cline{2-5}
& 0.10 & 267 ($6+(2)$) (62.5\%) & 406 ($4+(10)$) (50.0\%) & 615 ($6+(16)$) (45.45\%) \\ \cline{2-5}
& 0.15 & 587 ($10+(8)$) (66.67\%) & 891 ($6+(24)$) (56.67\%) & 1442 ($12+(38)$) (54.0\%) \\ \hline
\end{tabular}
\label{tab:Cand_Accuracy_McCarthy}
\end{small}
\end{table*}

\noindent{\bf Lau's method:} In Lau's method, consider the word `navigation'. Its news-specific sense cluster corresponds to a device to accurately ascertaining one's position and planning and following a route. The sense clusters in books corpus relate to navigation as a passage for ships among other senses and are different from the news-specific sense. The participants accordingly evaluated it as a news-specific sense. For the word `fencing', the book-specific cluster corresponds to the sense of fencing as a sports in which participants fight with swords under some rules. We can see that the first sense cluster from news corpus has a similar sense and accordingly, it was not judged as a corpus-specific sense. Finally, the book-specific cluster of `stalemate' does not denote any coherent sense, as also judged by the evaluators.

\noindent{\bf McCarthy's method:} In McCarthy's method, consider the word `pisces'. The book-specific cluster corresponds to the  12$^\textrm{th}$ sign of the zodiac in astrology. None of the clusters in the news corpus denote this sense and it was evaluated as book-specific. For the word `filibuster', the news-specific sense corresponds to an adventurer in a private military action in a foreign country. We can see that the cluster in the other corpus has the same sense and was not judged as corpus-specific. The news-specific sense cluster for the word `agora' does not correspond to any coherent sense of the word and was accordingly judged.

\noindent{\bf Mitra's method:} Finally, coming to Mitra's method, consider the word `chain'. Its news-specific cluster corresponds to the sense of a series of establishments, such as stores, theaters, or hotels, under a common ownership or management. The sense clusters in books corpus, on the other hand,  relate to chemical bonds, series of links of metals, polymers, etc. Thus, this sense of `chain' was evaluated as news-specific. Take the word `divider'. Its book-specific cluster corresponds to an electrical device used for various measurements. We can see that some of the clusters in the news corpus also have a similar sense (e.g., `pulses, amplifiers, proportional, pulse, signal, frequencies, amplifier, voltage'). Thus, this particular sense of `divider' was not judged as a corpus-specific sense. Finally, the news-specific cluster of 
the word `explanations' does not look very coherent and was judged as not representing a sense of explanations.

In general, corpus-specific senses, such as `navigation' as `gps, device, software' being news-specific, `pisces' as `12$^\textrm{th}$ sign of the zodiac' being book-specific and `chain' as `series of establishment' being news-specific look quite sensible.

\section{Parameter Tuning}
\label{sec:parameter}
To make our experiments more rigorous, we performed parameter tuning on Lau's and McCarthy's method to find the optimal accuracy value. We decided to select 50 words from each method to evaluate. 11 words out of these are from the time period 1987--1995 and the rest from the time period 2006--2008.

\noindent{\bf Lau's method:} For Lau's method, the thresholds represent maximum similarity. So, a lower value will be more restrictive as compared to a higher value. We selected three thresholds (0.30, 0.35, 0.40) for Lau's method for our experiment. Table \ref{tab:Accuracy_Lau} shows the total number of candidate words, words selected and average accuracy (overall confidence) of each threshold. First, we randomly selected 0.26\% words from the most restrictive threshold (i.e., 0.30). For the next threshold (0.35), since it contains all the words of the lower threshold (0.30), we we randomly selected 0.26\% words from the remaining 3715 words. We did the same for the threshold 0.40 again. Using the 50 words thus obtained, we performed the evaluation. We used the same evaluation method as outlined in Section~\ref{sec:evaluation}.



\noindent{\bf McCarthy's method:} For McCarthy's method, we have an upper and a lower threshold. A higher value for upper threshold and/or a lower value for lower threshold, would mean that it is more restrictive. Thus, a value of 0.45 for upper threshold and 0.05 for lower threshold would be the most restrictive in our set of thresholds. The total number of words, the number of words selected for evaluation and overall confidence are shown in Table~\ref{tab:Cand_Accuracy_McCarthy}. We used the same technique as we applied for Lau's method to evaluate a total of 50 words. 

We can see that a higher value (less restrictive) of the threshold provides better results in case of Lau. For McCarthy, we infer that a higher value (more restrictive) of upper threshold and a higher value (less restrictive) of the lower threshold is optimal.
\begin{table}[thb]
\centering
\begin{small}
\caption{Average accuracy for different threshold values in Lau's method.}
\begin{tabular}{|c|c|c|c|}
\hline
   Threshold & 0.30 & 0.35 & 0.40 \\ \hline
   Total Words & 11537 & 15252 & 19745 \\ \hline
   Words Selected & 30 & $9+(30)$ & $11+(39)$ \\ \hline
Average     & 16.67\%       & 28.2\% & 32.0 \%      \\ \hline
\end{tabular}
\label{tab:Accuracy_Lau}
\end{small}
\end{table}

\section{Conclusions and future work}
\label{sec:conclusion}
To summarize, we adapted three different methods for novel and predominant sense detection to identify cross corpus-specific word senses. In particular, we used 
multi-stage filtering to restrict the candidate senses by Mitra's method, used JS similarity across the sense clusters of two different corpora sources in Lau's method and used thresholds on the normalized prevalence score as well as the concept of denoting sense cluster using the most contributing neighbors in McCarthy's method. From the example cases, it is quite clear that after our adaptations, the outputs of the three proposed methods have very similar formats. Manual evaluation results were quite decent and in most of the cases, overall confidence in the methods was around 45-60\%. There is certainly scope in future for using advanced methods for comparing sense clusters, which can improve the accuracy of discriminative sense detection by these algorithms. Further, it will also be interesting to look into novel ways of combining results from different approaches.

\end{document}